\documentclass[10pt,twocolumn,letterpaper]{article}

\usepackage{cvpr}
\usepackage{times}
\usepackage{epsfig}
\usepackage{graphicx}
\usepackage{subfigure}
\usepackage{multirow}
\usepackage{amsmath}
\usepackage{amssymb}

  \newcommand{\bx}{\mathbf{x}}
  \newcommand{\by}{\mathbf{y}}
  \newcommand{\bw}{\mathbf{w}}
  \newcommand{\bg}{\mathbf{g}}
  \newcommand{\bh}{{\mathbf{h}}}
  \newcommand{\bc}{{\mathbf{c}}}
  \newcommand{\br}{{\mathbf{r}}}
  \newcommand{\bs}{{\mathbf{s}}}
  \newcommand{\bb}{{\mathbf{b}}}
  \newcommand{\bv}{{\mathbf{v}}}
  \newcommand{\bW}{{\mathbf{W}}}
  \newcommand{\bV}{{\mathbf{V}}}
  \newcommand{\bd}{{\mathbf{d}}}

\makeatletter
\newcommand*{\rom}[1]{\expandafter\@slowromancap\romannumeral #1@}
\makeatother

\usepackage[pagebackref=true,breaklinks=true,letterpaper=true,colorlinks,bookmarks=false]{hyperref}

 \cvprfinalcopy % *** Uncomment this line for the final submission

\def\cvprPaperID{1541} % *** Enter the CVPR Paper ID here

\begin{document}

%%%%%%%%% TITLE
\title{Hierarchical Attention-Based Recurrent Highway Networks\\ for Time Series Prediction}

%\author{Yunzhe Tao$^{1}$\thanks{Work performed when the author was with Tencent AI Lab.}, Lin Ma$^2$, Jian Liu$^3$, Wei Liu$^2$, Qiang Du$^1$\\
%$^1$Department of Applied Physics and Applied Mathematics, Columbia University, New York, NY 10027\\
%$^2$Tencent AI Lab, Shenzhen, China\\
%$^3$Tencent ..., Shenzhen, China\\
%y.tao@columbia.edu, forest.linma@gmail.com, albertjliu@tencent.com,\\wliu@ee.columbia.edu, qd2125@columbia.edu
%}

\author{Yunzhe Tao\\
Columbia University\\
New York, NY 10027\\
{\tt\small y.tao@columbia.edu}
\and
Lin Ma\\
Tencent AI Lab\\
Shenzhen, China\\
{\tt\small forest.linma@gmail.com}
\and
Weizhong Zhang\\
Tencent AI Lab\\
Shenzhen, China\\
{\tt\small zhangweizhongzju@gmail.com}
\and
Jian Liu\\
Tencent\\
Shenzhen, China\\
{\tt\small albertjliu@tencent.com}
\and
Wei Liu\\
Tencent AI Lab\\
Shenzhen, China\\
{\tt\small wl2223@columbia.edu}
\and
Qiang Du\\
Columbia University\\
New York, NY 10027\\
{\tt\small qd2125@columbia.edu}
}

\maketitle

\begin{abstract}
Time series prediction has been studied in a variety of domains. However, it is still challenging to predict future series given historical observations and past exogenous data. Existing methods either fail to consider the interactions among different components of exogenous variables which may affect the prediction accuracy, or cannot model the correlations between exogenous data and target data. Besides, the inherent temporal dynamics of exogenous data are also related to the target series prediction, and thus should be considered as well. To address these issues, we propose an end-to-end deep learning model, {\em i.e.,} Hierarchical attention-based Recurrent Highway Network (HRHN), which incorporates spatio-temporal feature extraction of exogenous variables and temporal dynamics modeling of target variables into a single framework. Moreover, by introducing the hierarchical attention mechanism, HRHN can adaptively select the relevant exogenous features in different semantic levels. We carry out comprehensive empirical evaluations with various methods over several datasets, and show that HRHN outperforms the state of the arts in time series prediction, especially in capturing sudden changes and sudden oscillations of time series.
\end{abstract}

% introduction
\section{Introduction}
\begin{figure}[!htb]
\centering
\subfigure[]{
        \includegraphics[width=.48\linewidth, height=.16\textheight]{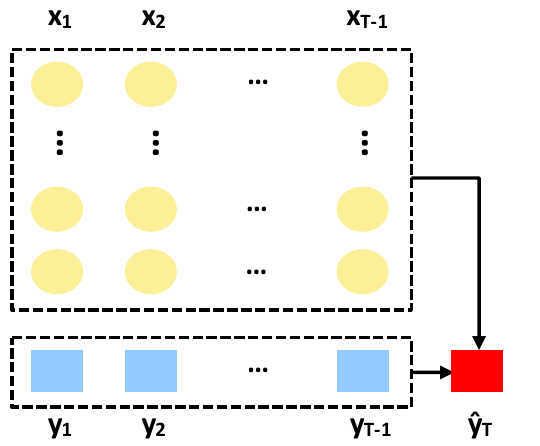}}
\subfigure[]{
        \includegraphics[width=.48\linewidth, height=.16\textheight]{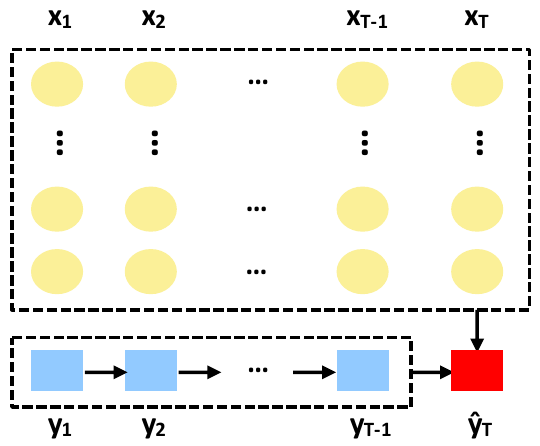}}\\
\subfigure[]{
        \includegraphics[width=.48\linewidth, height=.16\textheight]{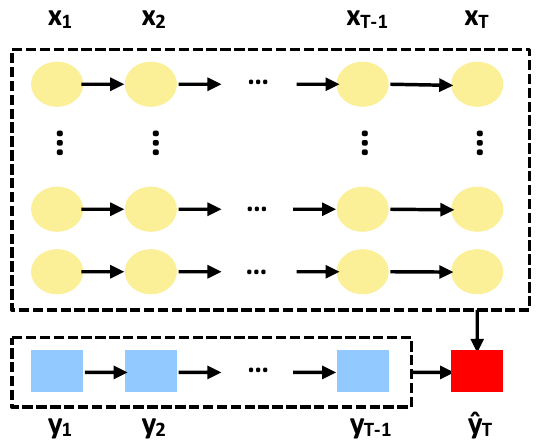}}
\subfigure[]{
        \includegraphics[width=.48\linewidth, height=.157\textheight]{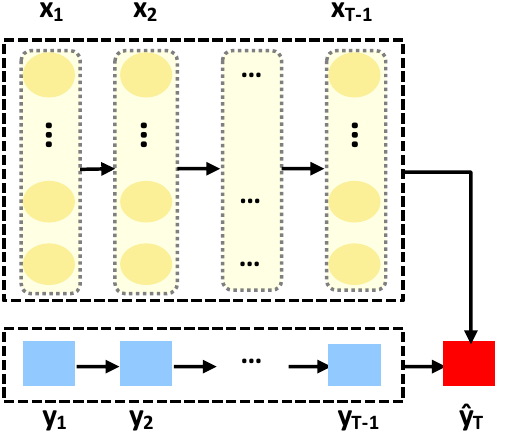}}
      \caption{Comparison of different dependencies in predicting $\hat{\by}_T$. (a) $\hat{\by}_T$ is linearly related to the historical observations and exogenous data. (b) $\hat{\by}_T$ is related to the temporal dynamics of historical observations and all the exogenous data (from the past and at time $T$). (c) $\hat{\by}_T$ is related to the historical observations and all the exogenous data as well as their temporal dynamics. (d) $\hat{\by}_T$ is related to the historical observations and exogenous data as well as their spatio-temporal dynamics.}
\label{fig:task}
\end{figure}

Time series modeling, as an example of signal processing problems, has played an important role in a variety of domains, such as complex dynamical system analysis \cite{dyn_sys}, speech analysis \cite{speech}, noise filtering \cite{noise}, and financial market analysis \cite{fin}. It is of interest to extract temporal information, uncover the correlations among past observations, analyze the dynamic properties, and predict future behaviors. Among all these application scenarios, the future behavior prediction task is of the greatest interest \cite{pred1,pred2}. However, such task is usually challenging due to non-stationarity, non-linearity, small sample size, and high noise of time series data.

The goal of time series forecasting is to generate the future series $\hat{\by}_T$ based on the historical observations $\by_1$, $\by_2,\cdots,\by_{T-1}$. Besides, the observations $\by_t$ are often related to some exogenous variables $\bx_t$. Different models have been proposed for time series prediction with access to the exogenous data. For example, as illustrated in Fig.~\ref{fig:task}(a), the autoregressive moving-average model with exogenous inputs (ARMAX) assumes that $\hat{\by}_T$ relies on not only the historical observations but also the past exogenous variables $\bx_1$, $\bx_2,\cdots,\bx_{T-1}$. However, this method assumes that the underlying model is linear, which limits its applications to real-world time series. As shown in Fig.~\ref{fig:task}(b), the mixed history recurrent neural network (MIST-RNN) model \cite{mistrnn} includes both past exogenous data and $\bx_T$, thus making more precise predictions. One problem is that MIST-RNN treats exogenous variables indistinguishably, ignoring their inherent temporal dynamics. Recently, the Dual-Attention Recurrent Neural Network (DA-RNN) model \cite{DA-RNN} was proposed to exploit the temporal dynamics of exogenous data in predicting $\hat{\by}_T$ (given in Fig.~\ref{fig:task}(c)). Although DA-RNN achieves better performance in some experiments, it is doubtful whether this method can be widely used in practice. Because in time series prediction of a future time $T$, the exogenous data $\bx_T$ is in general unavailable. Moreover, DA-RNN does not consider the correlations among different components of exogenous data, which may lead to poor predictions of complex real-world patterns. Therefore, how to make reliable future predictions based solely on the past exogenous data and historical observations still remains as an open question.

To address the aforementioned issues, as illustrated in Fig.~\ref{fig:task}(d), we need to first model the ``spatial'' (we use this word for convenience in opposite to ``temporal'') relationships between different components of the exogenous data $\bx_t$ at each time step, which usually present strong correlations with the observation $\by_t$. Second, we need to model the temporal behaviors of the historical observations and exogenous series as well as their interactions. In addition, the temporal information is usually complicated and may occur at different semantic levels. Therefore, how to fully exploit the spatial and temporal properties of the historical observations and exogenous data are two key problems for time series prediction.

In this paper, we propose an end-to-end neural network architecture, {\em i.e.}, Hierarchical attention-based Recurrent Highway Network (HRHN), to forecast future time series based on the observations and exogenous data only from the past. The contributions of this work are three-fold:
\begin{itemize}
\item We use a convolutional neural network (ConvNet) to learn the spatial interactions among different components of exogenous data, and employ a recurrent highway network (RHN) to summarize the exogenous data into different semantics at different levels in order to fully exploit and model their temporal dynamics.
\item We propose a hierarchical attention mechanism, performing on the discovered semantics at different levels, for selecting relevant information in the prediction.
\item Extensive experimental results on three different datasets demonstrate that the proposed HRHN model can not only yield high accuracy for time series prediction, but also capture sudden changes and oscillations of time series. 
%\item Extensive experimental results on three different datasets demonstrate that the proposed HRHN model can yield high accuracy for time series prediction, which outperforms the state-of-the-art methods. In particular, we carry out a video mining task on predicting the video views of a television series. 
\end{itemize}

%The rest of this paper is organized as follows. In Section 2 we briefly introduce the related work. Formal formulation of the problem considered in this paper is presented in Section 3. Then in Section 4, we introduce the details of our proposed HRHN. Section 5 shows an extensive empirical evaluation on time series prediction over various methods and datasets. Finally, we offer conclusions and directions for future work in Section 6.

\begin{figure*}[!tb]
\centering
 \includegraphics[width=\linewidth]{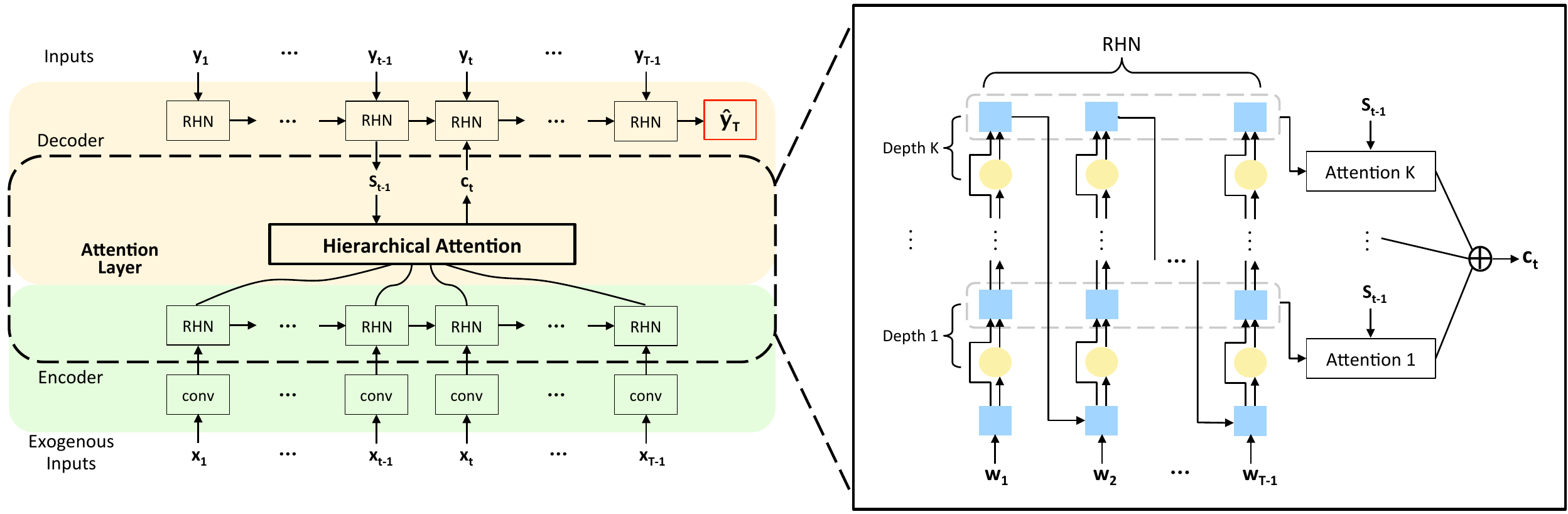}
 \caption{A graphical illustration of HRHN. In the encoder, a ConvNet extracts the ``spatial'' information of the exogenous inputs $\bx_1,\bx_2,\cdots,\bx_{T-1}$. Then an RHN learns the temporal dynamics from the representation vectors. Using a hierarchical attention mechanism, the decoder selects the most relevant spatio-temporal features of exogenous data and leverages another RHN to capture the long-term dependencies of target series $(\by_1,\by_2,\cdots,\by_{T-1})$ and produce the future prediction $\hat{\by}_T$. The bottom box is an illustration of the hierarchical attention approach. The encoder RHN reads the convolved features $(\bw_1,\bw_2,\cdots,\bw_{T-1})$ and models their temporal dependencies at different semantic levels. Then the hierarchical attention mechanism computes the soft alignment of hidden states in each layer. The context vector $\mathbf{c}_t$ that feeds into the decoder RHN is obtained by concatenating all attentions.}
 \label{fig:model}
 \end{figure*}

% related work
\section{Related Work}
In recent years, time series forecasting has been intensively studied. Among all the classical models, ARMA \cite{arma} has gained its popularity, which includes dynamic autoregressive and moving-average components. Moreover, the ARMAX model includes exogenous variables to model dynamics in historical observations. The NARMAX (nonlinear ARMAX) model is an extension of the linear ARMAX model, which represents the system via a nonlinear mapping from past inputs, outputs, and independent noisy terms to future outputs. However, these approaches usually use a pre-defined mapping and may not be able to capture the true underlying dynamics of time series.

RNNs \cite{rnn2,rnn} are a class of neural networks that are naturally suited for modeling time series data \cite{rnn4,rnn5}. The NARX-RNN \cite{narxrnn} model combines the capability of capturing nonlinear relationships by RNN with the effectiveness of gradient learning by NARX. However, traditional RNNs suffer from the issues of gradient vanishing and error propagation when learning long-term dependencies \cite{rnn3}. Therefore, special gating mechanisms that control access to memory cells have been developed, such as Long Short-Term Memory (LSTM) \cite{lstm} and Gated Recurrent Unit (GRU) \cite{gru}, which have already been used to perform time series predictions \cite{crnn1,crnn,crnn2}. Moreover, the MIST-RNN \cite{mistrnn} model has been developed based on a gating mechanism that is similar to GRU. This model makes the gradient decay's exponent closer to zero by using the mixed historical inputs. Although MIST-RNN can capture the long-term dependencies well, it does not consider the temporal behaviors of exogenous variables that may affect the dynamic properties of the observation data.

Recently, the encoder-decoder architectures \cite{enc-dec1,enc-dec} were developed for modeling sequential data. In time series forecasting, one usually makes prediction based on a long sequence of past observations, which forces the encoder to compress all the necessary information into a fixed-length vector, leading to poor performance as the input sequence length increases. Therefore, an attention mechanism \cite{attn} has been introduced, which improves the performance by learning a soft alignment between the input and output sequences. A representative work on time series prediction is the DA-RNN model \cite{DA-RNN}. In the encoder stage, DA-RNN exploits an input attention mechanism to adaptively extract relevant input features at each time step by referring to the previous encoder's hidden states. In the decoder stage, DA-RNN uses a classical attention mechanism to select relevant encoder's hidden states across all time steps. However, DA-RNN does not consider the ``spatial'' correlations among different components of exogenous data. More importantly, the classical attention mechanism cannot well model the complicated temporal dynamics, especially when the temporal dynamics may occur at different semantic levels.

% problem setup
\section{Problem Formulation}
A time series is defined as a sequence of real-valued observations with successive time stamps. In this paper, we focus on time series with identical interval lengths, and let $\by_t\in\mathbb{R}^d$ denote the observation measured at time $t$. Meanwhile, $\bx_t\in\mathbb{R}^n$ is the exogenous input at time $t$, which is assumed to be related to $\by_t$. 

Denote by $T$ the time window size. Our goal is to predict the current value of the target series $\hat{\by}_T$, given the historical observations $(\by_1,\by_2,\cdots,\by_{T-1})$ as well as the past (exogenous) input series $(\bx_1,\bx_2,\cdots,\bx_{T-1})$. More specifically, we aim to learn a nonlinear mapping $F(\cdot)$ such that
\begin{equation}
\hat{\by}_T = F(\by_1,\cdots,\by_{T-1},\bx_1,\cdots,\bx_{T-1})\,.
\end{equation}

% our model
\section{Hierarchical Attention-Based Recurrent Highway Networks}
We propose the Hierarchical attention-based Recurrent Highway Network (HRHN) for time series prediction. A detailed diagram of the system is plotted in Fig.~\ref{fig:model}. In the encoder, we first introduce the convolutional network (ConvNet) that can automatically learn the spatial correlations among different components of exogenous data. Then an RHN is used to model the temporal dependencies among convolved input features at different semantic levels. In the decoder, a novel hierarchical attention mechanism is proposed to select relevant encoded multi-level semantics. Another RHN is then introduced to capture the long-term temporal dependencies among historical observations, exploit the interactions between observations and exogenous data, and make the final prediction.

\subsection{Encoder}
The encoder takes historical exogenous data as inputs, and consists of a ConvNet for spatial correlation learning and an RHN for modeling exogenous temporal dynamics. 

\textbf{ConvNet for spatial correlations.} CNNs have already been widely applied to learning sequential data \cite{cnn3,cnn}. The key strength of CNN is that it automatically learns the feature representation by convolving the neighboring inputs and summarizing their interactions. Therefore, given the exogenous inputs $(\bx_1,\bx_2,\cdots,\bx_{T-1})$ with $\bx_t\in \mathbb{R}^n$, we apply independent local convolutions on each of the inputs to learn the interactions between different components of $\bx_t$. For some fixed $t$, assume that the number of convolutional layers is $L$ and the number of feature maps at the $\ell$-th layer is $F_\ell$. We also use kernels with fixed size of $1\times q$ for all layers. Then the convolution unit $i$ for feature map of type-$f$ at the $\ell$-th layer is given by
\begin{equation}\label{conv}
\bx_{(\ell,f)}^i = \gamma\left( \sum_{p=1}^{F_{\ell-1}} \sum_{j=0}^{q-1} \mathbf{k}_{(\ell,f)}\bx_{(\ell-1,p)}^{i+j} +\bb_{(\ell,f)}\right)\,,
\end{equation}
where $\mathbf{k}_{(\ell,f)}$ are the kernels for the type-$f$ feature map at the $\ell$-th layer, $\gamma$ is the activation function which is typically chosen to be ReLU \cite{relu}, and $\bb_{(\ell,f)}$ are the bias terms. Note that for the first layer ($\ell=1$), the inputs of \eqref{conv} are the exogenous inputs, namely $\bx_{(\ell-1,p)}^{i+j}=\bx_{(0,1)}^{i+j} = \bx_{i+j}$ since $F_0=1$.

A nonlinear subsampling approach, \textit{e.g.}, max pooling, is also performed between successive convolutional layers, which can reduce the size of feature maps so as to avoid overfitting and improve efficiency. Moreover, max pooling can remove the unreliable compositions generated during the convolution process. Assume that we adopt a $1\times s$ max-pooling process, which is given as
\begin{equation}
\bx_{(\ell+1,f)}^k = \max\left( \bx_{(\ell,f)}^{sk},  \bx_{(\ell,f)}^{sk+1}, \cdots, \bx_{(\ell,f)}^{sk+s-1} \right)\,,
\end{equation}
where $k$ starts from zero for clarity. 

After several layers of convolution and max-pooling, we feed the outputs to a fully connected layer, leading to a sequence of local feature vectors $(\bw_1,\bw_2,\cdots,\bw_{T-1})$ with $\bw_t\in \mathbb{R}^m$. Such a sequence can well exploit the interactions between different components at each time step.

\textbf{RHN for exogenous temporal dynamics.} Following the ConvNet, an RHN layer is used to model the temporal dynamics of exogenous series. Many sequential processing tasks require complex nonlinear transition functions from one step to the next. It is usually difficult to train gated RNNs such as LSTM and GRU when the networks go deeper. The RHN \cite{rhn} is designed to resolve such an issue by extending the LSTM architecture to allow step-to-step transition depth larger than one, which can capture the complicated temporal properties from different semantic levels. Let $\bg=G(\bw,\bW_G)$, $\br=R(\bw,\bW_R)$, and $\bc=C(\bw,\bW_C)$ be the outputs of nonlinear transformations $G,R$, and $C$, respectively. $R$ and $C$ typically utilize a sigmoid ($\sigma$) nonlinearity \cite{rhn} and are referred to as the \textit{transform} and \textit{carry} gates. Assume that in our model the RHN has recurrence depth of $K$ and $\bh_t^{[k]}\in\mathbb{R}^l$ is the intermediate output at time $t$ and depth $k$, where $t=1,2,\cdots,T-1$ and $k=1,2,\cdots,K$ with $\bh_t^{[0]}=\bh_{t-1}^{[K]}$. Moreover, let $\bW_{G,R,C}\in\mathbb{R}^{l\times m}$ and $\bV_{G_k,R_k,C_k}\in\mathbb{R}^{l\times l}$ represent the weight matrices of the $G$ nonlinearity and the $R$ and $C$ gates at the $k$-th layer, respectively. The biases are denoted by $\bb_{G_k,R_k,C_k}\in\mathbb{R}^l$. Then an RHN layer is described as
\begin{equation}\label{eqn:rhn}
\bh_t^{[k]} = \bg_t^{[k]} \cdot \br_t^{[k]} + \bh_t^{[k-1]} \cdot \bc_t^{[k]}\,,
\end{equation}
where
\begin{equation}
\begin{aligned}
& \bg_t^{[k]} = \text{tanh}\left( \bW_G \bw_t \mathbb{I}_{\{k=1\}}+\bV_{G_k}\bh_t^{[k-1]} + \bb_{G_k}\right)\,,\\
&\br_t^{[k]} = \quad\sigma \left( \bW_R \bw_t \mathbb{I}_{\{k=1\}}+\bV_{R_k}\bh_t^{[k-1]} + \bb_{R_k}\right)\,,\\
&\bc_t^{[k]} = \quad\sigma \left( \bW_C \bw_t \mathbb{I}_{\{k=1\}}+\bV_{C_k}\bh_t^{[k-1]} + \bb_{C_k}\right)\,,
\end{aligned}
\end{equation}
and $\mathbb{I}_{\{\}}$ is the indicator function meaning that $\bw_t$ is transformed only by the first highway layer. Moreover, at the first layer, $\bh_t^{[k-1]}$ is the RHN layer's output of the previous time step. The highway network described in \eqref{eqn:rhn} shows that the transform gate acts as selecting and controlling the information from history, and that the carry gate can carry the information between hidden states without any activation functions. Thus, the hidden states $\bh_t^{[k]}$ composed at different levels can capture the temporal dynamics of different semantics.

\begin{table*}[!tb]
\caption{The statistics of datasets.}
\[
 \begin{tabular}{|c|c|c|c|c|c|c|}
\hline \hline
\multirow{2}{*}{Dataset}&\,Exogenous\,& \, Target \, & \multicolumn{3}{|c|}{Size}  \\ \cline{4-6}
 &Data& Data&\,Training\,&\,Validation\,&\, \, Test \, \, \\ \hline \hline
 NASDAQ 100 Stock& 81 & 1& 35,100 & 2,730 & 2,730 \\ \hline
 Autonomous Driving & 6 & 4& 99,700 & 2,500 & 3,000 \\ \hline 
 \,Ode to Joy Video Views\, & 128 & 1& 38,600 & 2,270 & 4,560 \\ \hline \hline
\end{tabular} 
\]
 \label{table:data}
\end{table*}

\subsection{Decoder with Hierarchical Attention Mechanism}
In order to predict the future series $\hat{\by}_T\in\mathbb{R}^d$, we use another RHN to decode the input information which can capture not only the temporal dynamics of the historical observations but also the correlations among the observations and exogenous data. Besides, a hierarchical attention mechanism based on the discovered semantics at different levels is proposed to adaptively select the relevant exogenous features from the past. 

\textbf{Hierarchical attention mechanism.} Although the highway networks allow unimpeded information flow across layers, the information stored in different layers captures temporal dynamics at different levels and will thus have impact on predicting future behaviors of the target series. In order to fully exploit the multi-level representations as well as their interactions with the historical observations, the hierarchical attention mechanism computes the soft alignment of the encoder's hidden states $\bh_t^{[k]}$ in each layer based on the previous decoder layer's output $\bs_{t-1}:=\bs_{t-1}^{[K]}\in\mathbb{R}^p$. The attention weights of annotation at the $k$-th layer $\bh_i^{[k]}$  are given by
\begin{equation}
\alpha_{t,i}^{[k]} = \frac{e_{t,i}^{[k]}}{\sum_{j=1}^{T-1} e_{t,j}^{[k]}}\,,\quad 1\le i\le T-1\,,
\end{equation}
where
\begin{equation}
e_{t,i}^{[k]} = \bv^\text{T}_k\text{tanh}\left( \mathbf{T}_k \bs_{t-1}+ \mathbf{U}_k \bh_i^{[k]}\right)\,,
\end{equation}
and $\bv_k\in \mathbb{R}^l$, $\mathbf{T}_k\in\mathbb{R}^{l\times p}$ and $\mathbf{U}_k\in\mathbb{R}^{l\times l}$ are parameters to be learned. The bias terms have been omitted for clarity.  The \textit{alignment model} $e_{t,i}^{[k]}$ scores how well the inputs around position $i$ at the $k$-th layer match the output at position $t$ \cite{attn}. Then the soft alignment for layer $k$ is obtained by computing the sub-context vector $\bd_t^{[k]}$ as a weighted sum of all the encoder's hidden states in the $k$-th layer, namely
\begin{equation}
\bd_t^{[k]} = \sum_{i=1}^{T-1} \alpha_{t,i}^{[k]} \bh_i^{[k]}\,.
\end{equation}
At last, the context vector $\bd_t\in\mathbb{R}^{Kl}$ that we feed to the decoder is given through concatenating all the sub-context vectors in different layers, that is
\begin{equation}
\bd_t = \left[  \bd_t^{[1]}; \bd_t^{[2]};\cdots; \bd_t^{[K]}   \right]\,,\quad 1\le t\le T\,.
\end{equation}
Note that the context vector $\bd_t$ is time-dependent, which selects the most important encoder information in each decoding time step. Moreover, by concatenating the sub-context vectors, $\bd_t$ can also select the significant semantics from different levels of the encoder RHN, which encourages more interactions between the observations and exogenous data than the classical attention mechanism does. Once we obtain the concatenated context vector $\bd_t$, we can combine them with the given decoder inputs $(\by_1,\by_2,\cdots$ $,\by_{T-1})$, namely
\begin{equation}
\tilde{\by}_t = \tilde{\bW}  \by_t + \tilde{\bV} \bd_t  + \tilde{\bb}\,,\quad 1\le t\le T-1\,,
\end{equation}
where $\tilde{\bW}\in\mathbb{R}^{d\times d}$, $\tilde{\bV}\in\mathbb{R}^{d\times Kl}$ and $\tilde{\bb}\in\mathbb{R}^d$ are the weight matrices and biases to be learned. The time-dependent $\tilde{\by}_t$ represent the interactions between $\by_t$ and $\bd_t$, and are now the inputs of the decoder RHN layer. 

\textbf{RHN for target temporal dynamics.} Assume for simplicity that the decoder RHN also has recurrence depth of $K$. Then the update of the decoder's hidden states is given by
\begin{equation}
\bs_t^{[k]} = \tilde{\bg}_t^{[k]} \cdot \tilde{\br}_t^{[k]} + \bs_t^{[k-1]} \cdot \tilde{\bc}_t^{[k]}\,,
\end{equation}
where
\begin{equation}
\begin{aligned}
& \tilde{\bg}_t^{[k]} = \text{tanh}\left( \tilde{\bW}_G \tilde{\by}_t \mathbb{I}_{\{k=1\}}+\tilde{\bV}_{G_k}\bs_t^{[k-1]} + \tilde{\bb}_{G_k}\right)\,,\\
&\tilde{\br}_t^{[k]} = \quad\sigma \left( \tilde{\bW}_R \tilde{\by}_t \mathbb{I}_{\{k=1\}}+\tilde{\bV}_{R_k}\bs_t^{[k-1]} + \tilde{\bb}_{R_k}\right)\,,\\
&\tilde{\bc}_t^{[k]} = \quad\sigma \left( \tilde{\bW}_C \tilde{\by}_t \mathbb{I}_{\{k=1\}}+\tilde{\bV}_{C_k}\bs_t^{[k-1]} + \tilde{\bb}_{C_k}\right)\,.
\end{aligned}
\end{equation}
Here $\tilde{\bW}_{G,R,C}\in\mathbb{R}^{p\times d}$ and $\tilde{\mathbf{V}}_{G_k, R_k,C_k}\in\mathbb{R}^{p\times p}$ represent the weight matrices of the $G$ nonlinearity and the $R$ and $C$ gate functions, respectively, and $\tilde{\bb}_{G_k, R_k,C_k}\in\mathbb{R}^p$ are the bias terms to be learned.

As mentioned before, out goal is to find a nonlinear mapping such that 
\begin{equation}
\hat{\by}_T = F(\by_1,\cdots,\by_{T-1},\bx_1,\cdots,\bx_{T-1})\,.
\end{equation}
In our model, the prediction $\hat{\by}_T$ can be obtained by
\begin{equation}
\hat{\by}_T = \bW \bs_{T-1}^{[K]} + \bV \bd_T + \bb\,,
\end{equation}
where $\bs_{T-1}^{[K]}$ is the last layer's output and $\bd_T$ is its associated context vector. The parameters $\bW\in\mathbb{R}^{d\times p}$, $\bV\in\mathbb{R}^{d\times Kl}$ and $\bb\in\mathbb{R}^d$ characterize the linear dependency and produce the final prediction result.

% experiments
\section{Experiments}
In this section, we first introduce the datasets and their setup that are of interests to us in time series prediction. Then the parameters and performance evaluation metrics used in this work will be presented. At last, we compare the proposed HRHN model against some other cutting-edge methods, explore the performance of the ConvNet and the hierarchical attention approach in HRHN, and study the parameter sensitivity.

\begin{table*}[!tb]
\caption{Time series prediction results over different datasets (best performance is displayed in \textbf{boldface} in each case).}
\[
 \begin{tabular}{|c|c|c|c|c|c|c|c|c|}
\hline \hline
\multirow{2}{*}{Model} & \multicolumn{3}{|c|}{NASDAQ 100 Stock} & \multicolumn{2}{|c|}{Autonomous Driving} &  \multicolumn{3}{|c|}{Ode to Joy Video Views}  \\  \cline{2-9}
 & \,RMSE\,  & \,  MAE  \,  &MAPE (\%)& \,\,\, RMSE \,\,\, & MAE  & \,RMSE\, & \,\, MAE \,\,  &MAPE (\%) \\  \hline \hline
 ARIMA &  1.447 & 0.914 & 0.0185  & 3.135 & 2.209  & 401.18 &271.53 & 1.251  \\ \hline
  LSTM &  1.426& 0.919 & 0.0186  & 0.876 & 0.571  &397.16 & 272.88 & 1.246 \\ \hline
  GRU  & 1.437 & 0.937  & 0.0190 & 0.877 & 0.579  &398.57 & 270.15 &1.236 \\ \hline
 \, DA-RNN\, & 1.418 & 0.910  &  0.0184 & 0.871 & 0.569  & 398.29&269.28 & 1.235\\ \hline
  HRHN & \bf{1.401} & \bf{0.894} & \bf{0.0177} & \bf{0.860} & \bf{0.563}  & \bf{388.53} & \bf{263.17} & \bf{1.208} \\ \hline \hline
\end{tabular} 
\]
 \label{table:result}
\end{table*}

\subsection{Datasets and Setup}
We use three datasets to test and compare the performance of different methods in time series prediction. The statistics of datasets are given in Table \ref{table:data}.

NASDAQ 100 Stock dataset \cite{DA-RNN} collects the stock prices of 81 major corporations and the index value of NASDAQ 100, which are used as the exogenous data and the target observations, respectively. The data covers the period from July 26, 2016 to December 22, 2016, 105 trading days in total, with the frequency of every minute. We follow the original experiment \cite{DA-RNN} and use the first 35,100 data points as the training set and the following 2,730 data points as the validation set. The last 2,730 data points are used as the test set.

In addition, we also consider a state-changing prediction task in the high-speed autonomous driving \cite{autodriving}. Our goal is to predict the state derivatives of an autonomous rally car. The vehicle states include roll angle, linear velocities ($x$ and $y$ directions) and heading rate, four dimensions in total. The exogenous measurements are vehicle states and controls including steering and throttle. The training set of 99,700 data points contains data from several different runs in one day with the frequency of every 30 seconds of high-speed driving. The validation and test sets are recorded as one continuous trajectory in the same day with sizes of 2,500 and 3,000, respectively.

In particular, a video mining task is considered in our work. We take the first two seasons of Ode to Joy television drama as the dataset which aired in 2016 and 2017, respectively. Our goal is to predict the accumulated video views (VV) given the historical observations as the target data and the features extracted from the video as the exogenous data. The VV data is crawled from a public video website with the frequency of every five seconds and the exogenous features are obtained from pre-training the video by Inception-v4 \cite{2017inception}. The video has a resolution of $1920\times1072$ and is sampled at 25 frames per second. We use Inception-v4 to first extract the features of each frame, which yields a sequence of 1536 dimensional representations. Then we take the average of frames for each second and stack every five vectors, which leads to a sequence of 7680 dimensional vectors, representing the features of the video at every five seconds. Before feeding them into HRHN, we adopt a one-layer dense convolutional network for dimensionality reduction, resulting in 128 dimensional exogenous variables as the inputs of the encoder.

\begin{figure*}[!tb]
\centering
 \includegraphics[width=\linewidth]{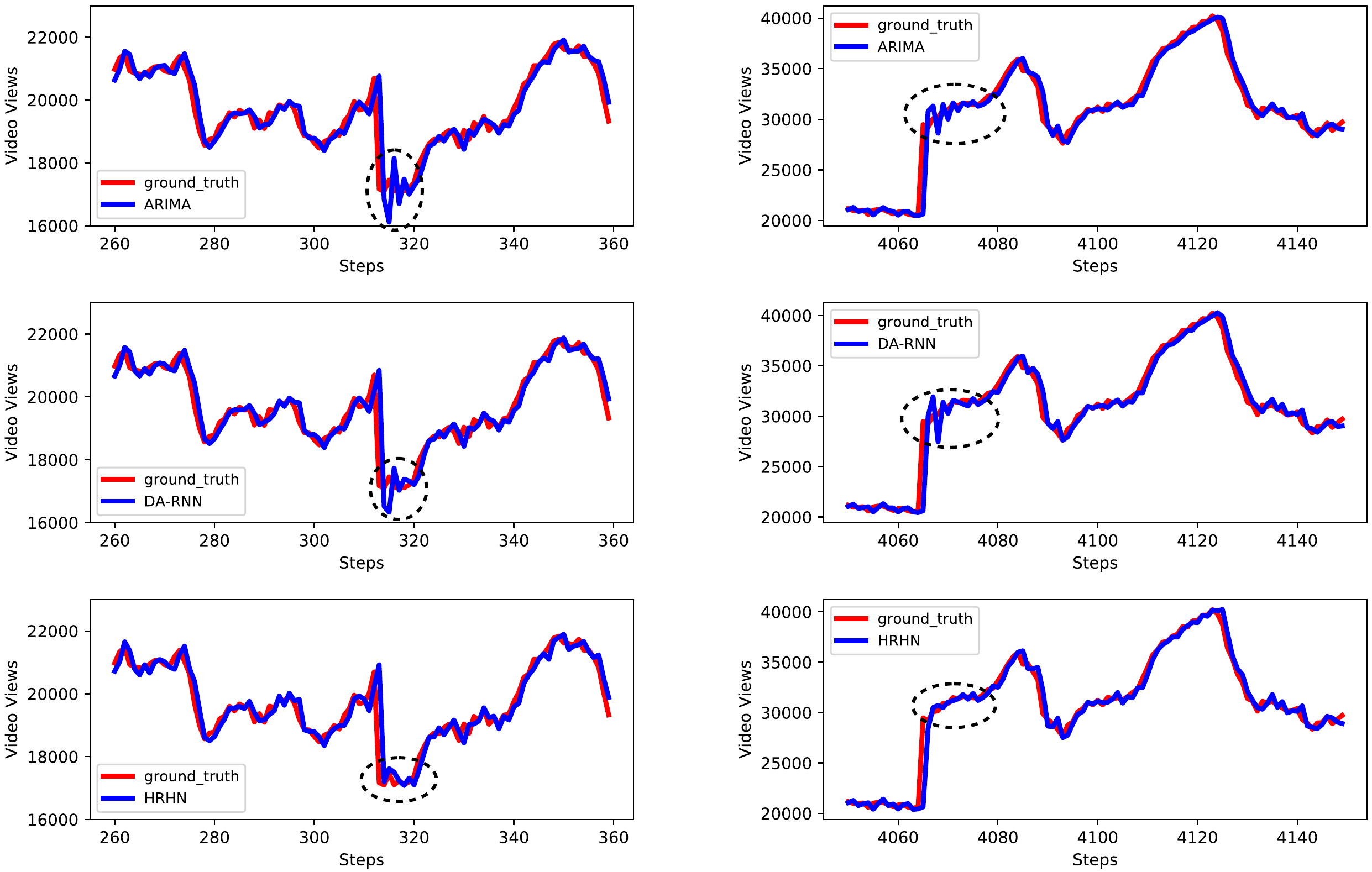}
     \caption{Visual comparison of HRHN (bottom) against ARIMA (top) and DA-RNN (middle) in predicting sudden changes on the video views dataset.}
      \label{fig:plot}
 \end{figure*}

\subsection{Parameters and Performance Evaluation}
For simplicity, we assume the RHN layer has same size in the encoder and the decoder of our model. Therefore, the parameters in our model include the number of time steps in the window $T$, the size of convolutional layers, kernels and max-pooling for the ConvNet, and the size of hidden states and recurrence depth for the RHN. To evaluate the performance of HRHN in different datasets, we choose different but fixed parameters. In NASDAQ 100 Stock dataset, we choose $T=11$, \textit{i.e.}, assuming that the target series is related to the past 10 steps, and the RHN layer with size of $128\times 2$, namely the RHN has the recurrence depth of 2 and 128 hidden states in each layer. For the ConvNet, we use three convolutional layers with kernel width of 3, and 16, 32 and 64 feature maps respectively, followed by a $1\times 3$ max-pooling layer after each convolutional layer. In the autonomous driving dataset, we use only one convolutional layer with $1\times 3$ kernels and 64 feature maps, followed by a $1\times 2$ max-pooling process. Besides, we let $T=11$ and RHN size to be $128\times 3$. At last for the Ode to Joy Video Views dataset, we choose $T=11$ and the size of RHN to be $64\times 4$. Again, only one layer of convolution and max-pooling are used with pooling size of $1\times 3$ and 128 feature maps. In order to fully convolve the features of every second, we use a kernel with width of 384, which is the dimension of stacking 3-second features. Moreover, we employ the strides with shape of $1\times 128$ to avoid involving redundant information.

In the training process, we adopt the mean squared error as objective function:
\begin{equation}
\mathcal{O}(\by_T,\hat{\by}_T) = \frac{1}{N} \sum_{i=1}^N \sum_{d=1}^D \left(\hat{\by}_T^i[d] - \by_T^i[d] \right)^2\,,
\end{equation}
where $N$ is the number of training samples and $D$ is the dimension of target data. All neural models are trained using the Adam optimizer \cite{adam} on a single NVIDIA Tesla GPU.
 
We consider three different metrics for performance evaluation on the single-step prediction task, namely the single-step error is measured as the root mean squared error (RMSE), the mean absolute error (MAE), and the mean absolute percentage error (MAPE) \cite{DA-RNN}. Note that since we have multivariate target data in the experiments, we define the metrics of multivariate data as the average of the metrics along all dimensions. More specifically, assume that $\by_T\in\mathbb{R}^D$ is the target value and $\hat{\by}_T$ is the predicted value at time $T$, then the RMSE is defined as 
\begin{equation}
\textbf{RMSE}=\frac{1}{D}\sum_{d=1}^D \sqrt{\frac{1}{N} \sum_{i=1}^N  \left(\hat{\by}_T^i[d] - \by_T^i[d] \right)^2}\,,
\end{equation}
and the MAE and MAPE are given by
\begin{equation}
\textbf{MAE}= \frac{1}{DN} \sum_{i=1}^N \sum_{d=1}^D \left| \hat{\by}_T^i[d] - \by_T^i[d] \right| \,,
\end{equation}
and
\begin{equation}
\textbf{MAPE}= \frac{1}{DN} \sum_{i=1}^N \sum_{d=1}^D \left| \frac{ \hat{\by}_T^i[d] - \by_T^i[d]}{\by_T^i[d]} \right| \,.
\end{equation}

\subsection{Result-\rom{1}: Time Series Prediction}
To demonstrate the effectiveness of HRHN, we first compare the performance of HRHN against some cutting-edge methods in the same prediction tasks. We choose ARIMA as the representative of the traditional methods and three other deep learning methods, including the attention-based encoder-decoder models with LSTM and GRU layers, and the DA-RNN model \cite{DA-RNN}. The prediction results over three datasets are presented in Table \ref{table:result}. For each dataset, we run each method 5 times and report the median in the table. Note that for autonomous driving dataset, there are many zero values after normalization, which leads to the issue in MAPE calculation. Therefore, the MAPE result is omitted on this dataset.

We can observe that the performance of deep learning methods based on neural networks is better than the traditional learning model (ARIMA), and the proposed HRHN achieves the best results for all the three metrics across all datasets, due to the following reasons. On one hand, the LSTM or GRU model only considers the temporal dynamics of exogenous inputs, while the DA-RNN model uses an input attention mechanism to extract relevant input features of exogenous data. However, DA-RNN does not capture the correlations among different components of the inputs. Our HRHN model can learn the spatial interactions of exogenous data by introducing the ConvNet. On the other hand, the RHN layer can well model the complicated temporal dynamics with different semantics at different levels, and the hierarchical attention mechanism can well exploit such information, which is better than the classical attention approach and other gating mechanisms. 

\begin{table*}[!tb]
\caption{Performance of different modules in HRHN.}
\[
 \begin{tabular}{|c|c|c|c|c|c|c|c|c|}
\hline \hline
\multirow{2}{*}{Model} & \multicolumn{3}{|c|}{NASDAQ 100 Stock} & \multicolumn{2}{|c|}{ Autonomous Driving} &  \multicolumn{3}{|c|}{Ode to Joy Video Views}  \\  \cline{2-9}
 & RMSE & \,MAE\,   & MAPE (\%) & \,\,\, RMSE \,\,\, &  MAE  & RMSE  &  \,MAE\,  & MAPE (\%)  \\  \hline \hline
  RHN & 1.408 & 0.899   & 0.0181  & 0.865 & 0.566  &395.51 & 269.41& 1.234\\ \hline 
 RHN + ConvNet & 1.405 & 0.896  & 0.0179 &0.864 & 0.565  &391.18 &266.20 & 1.222\\ \hline
  RHN + HA& 1.404 & 0.897 & 0.0180 & 0.864 & 0.564  & 393.55& 267.54& 1.226\\ \hline \hline
\end{tabular} 
\]
 \label{table:result2}
\end{table*}

\begin{table}[!tb] 
\caption{Performance of the attentions at different layers.} \vskip -1.2em
\[\begin{tabular}{|c|c|c|c|}
\hline \hline
\multirow{2}{*}{Model} & \multicolumn{3}{|c|}{Ode to Joy Video Views}\\ \cline{2-4}
&\, \,RMSE\, \,& \, \,MAE\, \, & MAPE (\%)  \\ \hline \hline
\,\, RHN-attn3 \,\, &396.38 & 272.39 & 1.244\\ \hline
RHN-attn2 & 395.93& 266.63 & 1.225 \\ \hline
RHN-attn1 & 394.41 & 265.09 & 1.217 \\ \hline \hline
\end{tabular}\]
 \label{table:attn}
\end{table}

Besides, one can also expect that the difference of results among the methods is bigger on more complicated dynamic systems. For example, in the autonomous driving dataset, the deep learning models perform much better than ARIMA due to the complexity of multi-variate target outputs. In addition, the difference among methods is bigger in the video views dataset than that in the stock dataset, since the patterns within a video is more complex than that in stock prices. 

In time series prediction, it is sometimes more interesting to compare the performance on capturing the so-called rare events, \textit{e.g.}, an oscillation after a stable growth or decay, or a huge sudden change during oscillations. The visual comparison on predicting such events by HRHN against ARIMA and DA-RNN is given in Fig.~\ref{fig:plot}. We plot some test samples and the corresponding prediction results over the Ode to Joy video views dataset. One can see that around steps 320 (left three circles) and 4070 (right three circles), where the oscillation occurs after a stable change, HRHN fits the ground truth better than others, which again illustrates the ability of ConvNet in summarizing the interactions of exogenous inputs and the hierarchical attention approach in selecting important multi-level features.

\subsection{Result-\rom{2}: Effectiveness of Modules in HRHN}
%\vskip -1em
The effectiveness of HRHN can also be shown via a step-by-step justification. We compare HRHN against the classical attention-based RHN encoder-decoder model and the setting that only employs the ConvNet (RHN + ConvNet) or the hierarchical attention mechanism (RHN + HA). The results are presented in Table \ref{table:result2}, where we again run each experiment 5 times and report the median. We provide a brief analysis as follows based on the results.

\textbf{RHN.} One can notice that the single RHN encoder-decoder model outperforms DA-RNN in all metrics and datasets except the MAE for video views dataset (which is also comparable). Although RHN cannot model the spatial correlations of exogenous inputs, it can well capture their temporal dynamics from different levels by allowing deeper step-to-step transitions.

\textbf{ConvNet.} From the results in Table \ref{table:result} and Table \ref{table:result2}, we can observe that the RHN equipped with ConvNet consistently outperforms DA-RNN and single RHN, which suggests that by convolving the neighboring exogenous inputs, ConvNet is able to summarize and exploit the interactions between different components at each time step, which has impacts on predicting target series.

\begin{figure}[!!tb]
\vskip .8em
\centering
\includegraphics[width=.49\linewidth]{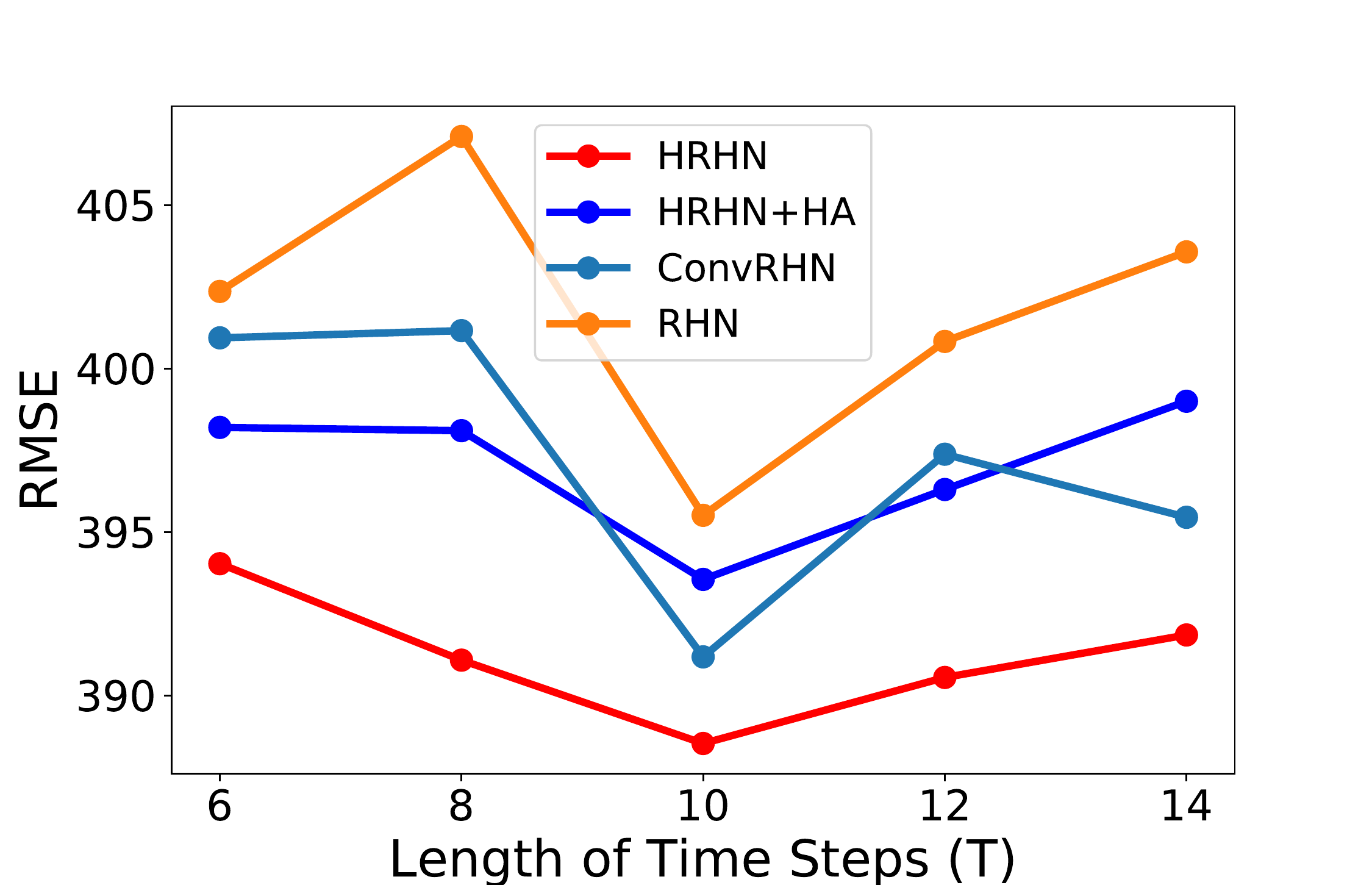}
\includegraphics[width=.49\linewidth]{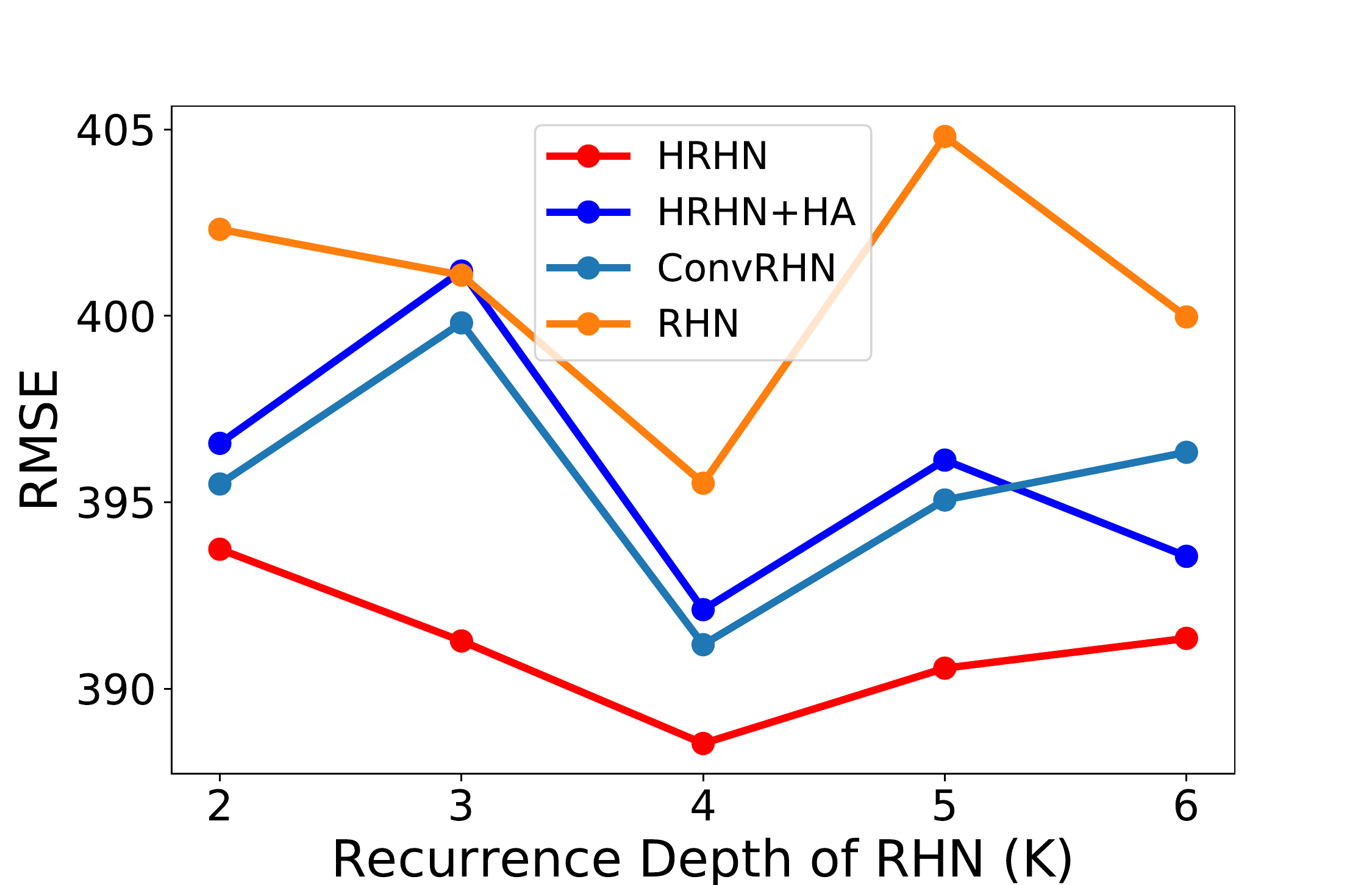}
 \caption{Parameter sensitivity of models over the Ode to Joy video views dataset. Left: RMSE {\em vs.} length of time steps $T$. Right: RMSE {\em vs.} recurrence depth of RHN.}
\label{fig:sens}
\end{figure}

\textbf{Hierarchical attention.} Similarly, the RHN equipped with the hierarchical attention also outperforms DA-RNN and single RHN. To further demonstrate the effectiveness of the hierarchical attention mechanism, we employ the classical attention-based RHN approach where the attention is computed from each single intermediate layer of the encoder RHN, and compare the performance with that of the aforementioned RHN model. We still take the video views dataset as an illustration since the largest number of recurrence depth is used in this task. Because a four-layer RHN is adopted in the prediction, we compare the results obtained from Table~\ref{table:result2} to the performance of the RHN with attentions of the three hidden layers. From the results stated in Table~\ref{table:attn}, we can confirm that useful information is also stored in the intermediate layers of RHN and the hierarchical attention mechanism can extract such information in predicting future series.

\subsection{Result-\rom{3}: Parameter Sensitivity}
At last, we can also study the parameter sensitivity of the proposed methods, especially the HRHN model. Parameters of interests include the length of time steps $T$ and the size of recurrence depth $K$ of the encoder and decoder RHN. The video views dataset is used again for demonstration. When we vary $T$ or $K$, we keep the other fixed. By setting $K=4$, we plot the RMSE against different lengths of time steps in the window $T$ in Fig.~\ref{fig:sens} (left) and by setting $T = 10$, we also plot the RMSE against different recurrence depths for RHN in Fig.~\ref{fig:sens} (right). We compare the sensitivity results of several aforementioned methods, including the single RHN encoder-decoder model, RHN + ConvNet, RHN + HA, and HRHN.

We notice that the results of both cases and all the models do not differ much for different parameters. In particular, the difference on HRHN is the smallest among the models (less than 2\%), which implies the robustness of our HRHN model. Moreover, we can also observe that HRHN performs worse when the window size or recurrence depth is too small or too large, since the former leads to lack of sufficient information for feature extraction while the latter produces redundant features for capturing the correct temporal dependencies.

% conclusion
\section{Conclusion and Future Work}
In this paper, we proposed Hierarchical attention-based Recurrent Highway Network (HRHN) for time series prediction. Based upon the modules in the proposed model, HRHN can not only extract and select the most relevant input features hierarchically, but also capture the long-term dependencies of the time series. The extensive experiments on various datasets demonstrate that our proposed HRHN advances the state-of-the-art methods in time series prediction. In addition, HRHN is good at capturing and forecasting the rare events, such as sudden changes and sudden oscillations.

In the future, we intend to apply HRHN to other time series prediction or signal processing tasks, \textit{e.g.}, multi-step prediction (sequence to sequence learning). Moreover, we believe that our model can also produce useful information for forecasting future exogenous variables. Besides, we will investigate the correlations among different components of target variables as well. In this work, we predict all the components of multi-variate target series simultaneously. However, predicting one component within the target series should help us generate the predictions on the others, which also requires more investigations.

\subsubsection*{Acknowledgments}
This work is supported in part by US NSF CCF-1740833, DMR-1534910 and DMS-1719699.

{\small
\bibliographystyle{ieee}
\bibliography{hrhn}

\begin{thebibliography}{10}\itemsep=-1pt

\bibitem{cnn3}
O.~Abdel-Hamid, L.~Deng, and D.~Yu.
\newblock Exploring convolutional neural network structures and optimization
  techniques for speech recognition.
\newblock {\em Interspeech}, pages 3366--3370, 2013.

\bibitem{attn}
D.~Bahdanau, K.~Cho, and Y.~Bengio.
\newblock Neural machine translation by jointly learning to align and
  translate.
\newblock {\em arXiv preprint arXiv:1409.0473}, 2014.

\bibitem{crnn1}
P.~Bashivan, I.~Rish, M.~Yeasin, and N.~Codella.
\newblock Learning representations from eeg with deep recurrent-convolutional
  neural networks.
\newblock {\em arXiv preprint arXiv:1511.06448}, 2015.

\bibitem{rnn3}
Y.~Bengio, P.~Simard, and P.~Frasconi.
\newblock Learning long-term dependencies with gradient descent is difficult.
\newblock {\em IEEE transactions on neural networks}, 5(2):157--166, 1994.

\bibitem{enc-dec1}
K.~Cho, B.~V. Merri{\"e}nboer, D.~Bahdanau, and Y.~Bengio.
\newblock On the properties of neural machine translation: Encoder-decoder
  approaches.
\newblock {\em arXiv preprint arXiv:1409.1259}, 2014.

\bibitem{enc-dec}
K.~Cho, B.~V. Merri{\"e}nboer, C.~Gulcehre, D.~Bahdanau, F.~Bougares,
  H.~Schwenk, and Y.~Bengio.
\newblock Learning phrase representations using rnn encoder-decoder for
  statistical machine translation.
\newblock {\em arXiv preprint arXiv:1406.1078}, 2014.

\bibitem{gru}
J.~Chung, C.~Gulcehre, K.~Cho, and Y.~Bengio.
\newblock Empirical evaluation of gated recurrent neural networks on sequence
  modeling.
\newblock {\em arXiv preprint arXiv:1412.3555}, 2014.

\bibitem{pred1}
J.~T. Connor, R.~D. Martin, and L.~E. Atlas.
\newblock Recurrent neural networks and robust time series prediction.
\newblock {\em IEEE transactions on neural networks}, 5.2:240--254, 1994.

\bibitem{relu}
G.~E. Dahl, T.~N. Sainath, and G.~E. Hinton.
\newblock Improving deep neural networks for lvcsr using rectified linear units
  and dropout.
\newblock {\em Acoustics, Speech and Signal Processing (ICASSP), 2013 IEEE
  International Conference on}, pages 8609--8613, 2013.

\bibitem{mistrnn}
R.~DiPietro, N.~Navab, and G.~D. Hager.
\newblock Revisiting narx recurrent neural networks for long-term dependencies.
\newblock {\em arXiv preprint arXiv:1702.07805}, 2017.

\bibitem{rnn2}
J.~L. Elman.
\newblock Distributed representations, simple recurrent networks, and
  grammatical structure.
\newblock {\em Machine learning}, 7.2-3:195--225, 1991.

\bibitem{noise}
J.~Gao, H.~Sultan, J.~Hu, and W.-W. Tung.
\newblock Denoising nonlinear time series by adaptive filtering and wavelet
  shrinkage: a comparison.
\newblock {\em IEEE signal processing letters}, 17(3):237--240, 2010.

\bibitem{rnn4}
A.~Graves, A.-r. Mohamed, and G.~E. Hinton.
\newblock Speech recognition with deep recurrent neural networks.
\newblock {\em Acoustics, speech and signal processing (icassp), 2013 ieee
  international conference on}, pages 6645--6649, 2013.

\bibitem{lstm}
S.~Hochreiter and J.~Schmidhuber.
\newblock Long short-term memory.
\newblock {\em Neural computation}, 9.8:1735--1780, 1997.

\bibitem{adam}
D.~Kingma and J.~Ba.
\newblock Adam: A method for stochastic optimization.
\newblock {\em arXiv preprint arXiv:1412.6980}, 2014.

\bibitem{cnn}
Y.~LeCun and Y.~Bengio.
\newblock Convolutional networks for images, speech, and time series.
\newblock {\em The handbook of brain theory and neural networks}, 3361(10),
  1995.

\bibitem{narxrnn}
T.~Lin, B.~Horne, P.~Tino, and C.~Giles.
\newblock Learning long-term dependencies in narx recurrent neural networks.
\newblock {\em IEEE Transactions on Neural Networks}, 7(6):1329--1338, 1996.

\bibitem{dyn_sys}
Z.~Liu and M.~Hauskrecht.
\newblock A regularized linear dynamical system framework for multivariate time
  series analysis.
\newblock pages 1798--1804, 2015.

\bibitem{autodriving}
Y.~Pan, X.~Yan, E.~A. Theodorou, and B.~Boots.
\newblock Prediction under uncertainty in sparse spectrum gaussian processes
  with applications to filtering and control.
\newblock {\em ICML}, pages 2760--2768, 2017.

\bibitem{crnn}
S.~C. Prasad and P.~Prasad.
\newblock Deep recurrent neural networks for time series prediction.
\newblock {\em arXiv preprint arXiv:1407.5949}, 2014.

\bibitem{DA-RNN}
Y.~Qin, D.~Song, H.~Cheng, W.~Cheng, G.~Jiang, and G.~Cottrell.
\newblock A dual-stage attention-based recurrent neural network for time series
  prediction.
\newblock {\em arXiv preprint arXiv:1704.02971}, 2017.

\bibitem{speech}
L.~Rabiner and R.~Schafer.
\newblock {\em Digital processing of speech signals}.
\newblock Prentice Hall, 1978.

\bibitem{rnn}
D.~E. Rumelhart, G.~E. Hinton, and R.~J. Williams.
\newblock Learning representations by back-propagating errors.
\newblock {\em Cognitive modeling}, 5.3:1, 1988.

\bibitem{crnn2}
X.~Shi, Z.~Chen, H.~Wang, D.-Y. Yeung, W.-K. Wong, and W.-c. Woo.
\newblock Convolutional lstm network: A machine learning approach for
  precipitation nowcasting.
\newblock {\em NIPS}, pages 802--810, 2015.

\bibitem{rnn5}
I.~Sutskever, J.~Martens, and G.~E. Hinton.
\newblock Generating text with recurrent neural networks.
\newblock {\em Proceedings of the 28th International Conference on Machine
  Learning (ICML-11)}, pages 1017--1024, 2011.

\bibitem{2017inception}
C.~Szegedy, S.~Ioffe, V.~Vanhoucke, and A.~A. Alemi.
\newblock Inception-v4, inception-resnet and the impact of residual connections
  on learning.
\newblock In {\em AAAI}, pages 4278--4284, 2017.

\bibitem{pred2}
T.~Van~Gestel, J.~Suykens, D.~Baestaens, A.~Lambrechts, G.~Lanckriet,
  B.~Vandaele, B.~De~Moor, and J.~Vandewalle.
\newblock Financial time series prediction using least squares support vector
  machines within the evidence framework.
\newblock {\em IEEE Transactions on Neural Networks}, 12(4):809--821, 2001.

\bibitem{arma}
P.~Whittle.
\newblock {\em Hypothesis Testing in Time Series Anal- ysis}.
\newblock PhD thesis, 1951.

\bibitem{fin}
Y.~Wu, J.~M. Hern{\'a}ndez-Lobato, and Z.~Ghahramani.
\newblock Dynamic covariance models for multivariate financial time series.
\newblock {\em ICML}, pages 558--566, 2013.

\bibitem{rhn}
J.~G. Zilly, R.~K. Srivastava, J.~Koutn{\'\i}k, and J.~Schmidhuber.
\newblock Recurrent highway networks.
\newblock {\em arXiv preprint arXiv:1607.03474}, 2016.

\end{thebibliography}
}

\end{document}